\documentclass[10pt,twocolumn,letterpaper]{article}

\usepackage{wacv}              

\usepackage{amsthm}
\usepackage{enumitem}
\usepackage{amsmath}
\usepackage{amssymb}
\usepackage{amsfonts}
\usepackage{mathtools}
\usepackage{float}
\usepackage{textcomp}
\usepackage{booktabs}
\usepackage{multirow}
\usepackage{array}
\usepackage{graphicx}
\usepackage{adjustbox}
\usepackage{dsfont}
\usepackage{algorithm}
\usepackage{algorithmic}
\usepackage{tcolorbox}
\usepackage{xcolor} 
\usepackage{rotating}
\usepackage{hyperref}

\theoremstyle{plain}               
\newtheorem{theorem}{Theorem}[section]  
\newtheorem{lemma}[theorem]{Lemma}      
\theoremstyle{definition}          
\newtheorem{definition}[theorem]{Definition}
\theoremstyle{remark}              
\newtheorem{remark}[theorem]{Remark}
\newtheorem{corollary}[theorem]{Corollary}

\definecolor{myred}{RGB}{255,0,0}

\usepackage{colortbl}
\usepackage{xcolor,colortbl}
\definecolor{lightblue}{RGB}{225,245,255}
\definecolor{lightred}{RGB}{255,233,233}

\title{AGENet: Adaptive Edge-aware Geodesic Distance Learning for Few-Shot Medical Image Segmentation}

\author{
Ziyuan Gao\\
University College London\\
 {\tt\small clairegao0930@gmail.com}
}

\begin{document}
\maketitle

\begin{abstract}

Medical image segmentation requires large annotated datasets, creating a significant bottleneck for clinical applications. 
While few-shot segmentation methods can learn from minimal examples, existing approaches demonstrate suboptimal performance in precise boundary delineation for medical images, particularly when anatomically similar regions appear without sufficient spatial context.
We propose AGENet (Adaptive Geodesic Edge-aware Network), a novel framework that incorporates spatial relationships through edge-aware geodesic distance learning. Our key insight is that medical structures follow predictable geometric patterns that can guide prototype extraction even with limited training data. Unlike methods relying on complex architectural components or heavy neural networks, our approach leverages computationally lightweight geometric modeling.
The framework combines three main components: (1) An edge-aware geodesic distance learning module that respects anatomical boundaries through iterative Fast Marching refinement, (2) adaptive prototype extraction that captures both global structure and local boundary details via spatially-weighted aggregation, and (3) adaptive parameter learning that automatically adjusts to different organ characteristics. Extensive experiments across diverse medical imaging datasets demonstrate improvements over state-of-the-art methods. Notably, our method reduces boundary errors compared to existing approaches while maintaining computational efficiency, making it highly suitable for clinical applications requiring precise segmentation with limited annotated data.

\end{abstract}

\section{Introduction}
\label{sec:Introduction}

Medical image segmentation is fundamental for clinical diagnosis and treatment planning, but training effective segmentation models traditionally requires thousands of meticulously annotated images~\cite{Ronneberger15}. This extensive data requirement creates significant bottlenecks in medical settings where annotated data is inherently scarce due to high expert annotation costs, rare pathological conditions, and privacy constraints. Few-shot segmentation offers a promising solution by enabling models to learn from minimal support examples, making it particularly valuable for medical applications with limited training data availability~\cite{Snell17}. However, existing few-shot methods often struggle to capture the inherent geometric regularity and crucial edge sensitivity necessary for precise organ delineation in medical images. This limitation frequently leads to misclassifications of anatomically similar regions, exacerbated by insufficient training data~\cite{Snell17}.

\begin{figure}[t]
  \centering
    \includegraphics[width=\linewidth]{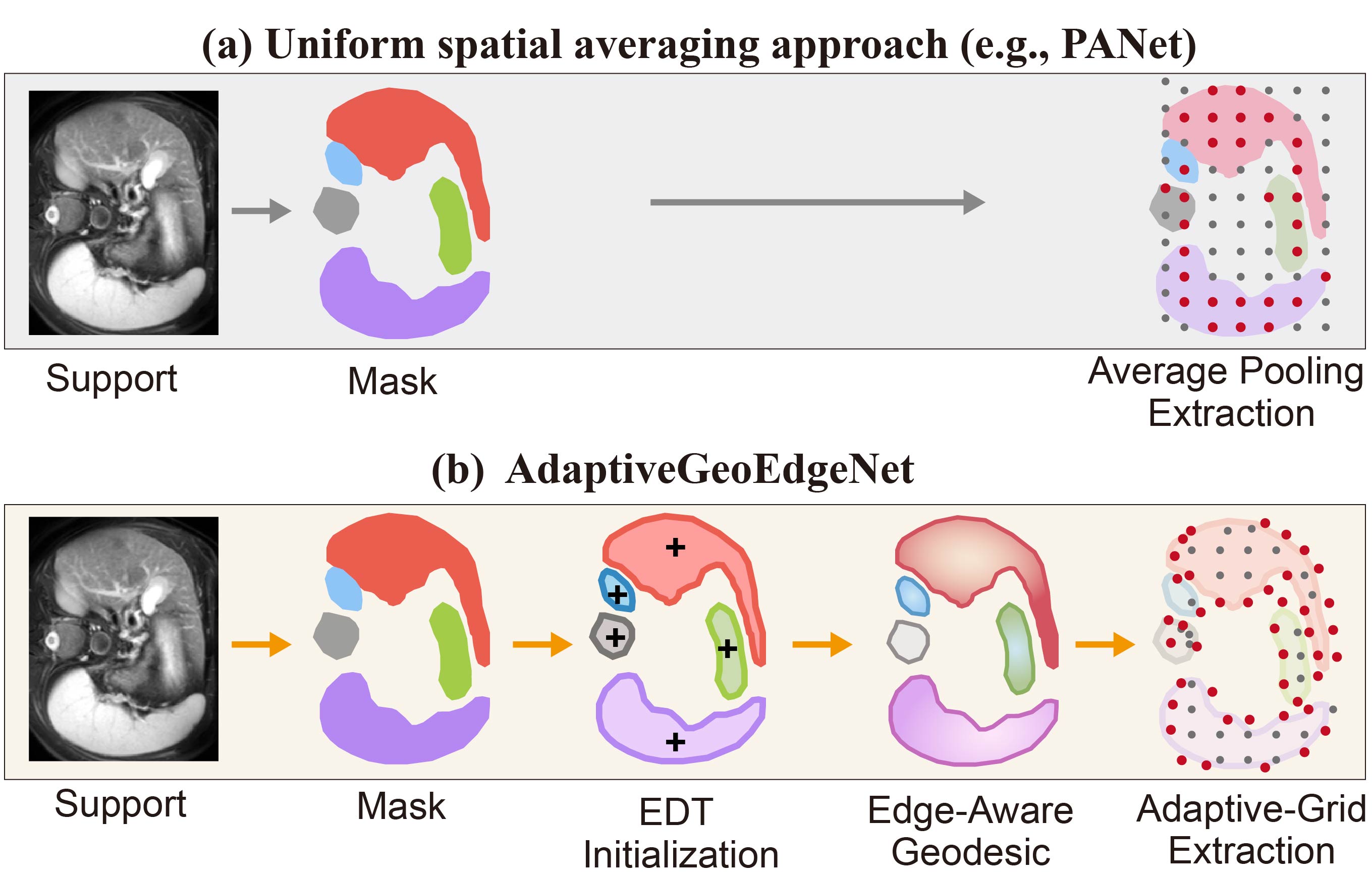}
    \caption{This figure compares two prototype extraction methods: (a) Conventional prototype-based approach (e.g., PANet) that relies on uniform spatial averaging to extract a single global prototype from the support mask.  (b) Our method employs edge-aware geodesic distance learning with adaptive multi-scale extraction.}
    \label{fig:concept}
\end{figure}

A key limitation in few-shot medical image segmentation stem lies in their spatial modeling problem. Existing prototype aggregation methods, including prototype-based networks~\cite{Snell17,PANet} and metric learning approaches~\cite{Vinyals16,Finn17}, treat all spatial locations within anatomical structures uniformly during prototype extraction. This uniform spatial averaging inadequately captures the anisotropic nature of medical structures, where pixels near organ boundaries contain diagnostically crucial information for tumor detection and surgical planning~\cite{Rosenthal18,vandenBerg18}, while interior pixels often represent homogeneous tissue regions with limited diagnostic significance~\cite{Schmidt93,Wazer93}. 

This limited spatial modeling problem manifests in two critical ways. First, it produces spatially-averaged prototypes that combine representative organ tissue with boundary transitional features, as pixels within organ masks receive insufficient differentiation based on their anatomical significance. Second, during query segmentation, these mixed prototypes can lead to similarity computation errors at organ boundaries, resulting in reduced segmentation accuracy where precision is most critical for clinical decision-making.
While recent advancements, including SSL-ALPNet~\cite{Ouyang} and DSPNet~\cite{Tang2024}, have demonstrated improvements through enhanced prototype extraction mechanisms, they primarily focus on feature-level refinements rather than addressing the fundamental spatial modeling challenges. These approaches do not explicitly incorporate geometric constraints or spatial priors, leaving a gap between feature-based prototype learning and anatomically-informed spatial understanding necessary for precise medical segmentation.

To overcome these challenges, we propose AGENet (Adaptive Geodesic Edge-aware Network), a novel edge-aware geodesic distance learning framework, as shown in Figure~\ref{fig:concept}. Instead of complex architectures or heavy neural networks, our approach transforms prototype extraction through computationally lightweight geometric modeling. This allows us to encode the spatial importance of different regions within medical structures, resulting in more effective prototype learning even with minimal supervision.

Our contributions are three-fold:
(1) We introduce an \textbf{edge-aware geodesic distance learning module}. This module combines foundational Euclidean Distance Transform (EDT) initialization, iterative Fast Marching refinement for geometric constraints, and adaptive parameter learning for domain-specific enhancement. This enables the computation of anatomically-coherent spatial importance maps that capture tissue boundaries.
(2) We develop an \textbf{adaptive prototype extraction framework} that captures both global anatomical structure and local boundary details. By integrating geodesic-weighted global prototypes with adaptive grid sampling, this hierarchical framework produces more discriminative representations for anatomical segmentation. This overcomes the spatial limitations of single-prototype methods, leading to more accurate results.
(3) We provide comprehensive evaluation and analysis across nine datasets (ACDC, ABD-MRI, ABD-CT, PH2, CVC-300, CHASE-DB1, Drishti-GS, DCA1) representing diverse data scarcity scenarios and anatomical complexities, with thorough ablation studies examining individual component contributions and computational efficiency analysis.
Experimental results demonstrate substantial improvements over state-of-the-art methods, achieving 79.56\% (1-shot) and 81.67\% (5-shot) mean Dice scores with 11.16mm and 8.39mm Hausdorff Distance, while maintaining computational efficiency competitive with existing approaches.

\section{Related Work}
\label{sec:Related_Work}

\paragraph{Prototype-Based Few-Shot Medical Segmentation}

Recent advancements in few-shot medical image segmentation have largely focused on enhancing medical-specific prototypes. SSL-ALPNet~\cite{Ouyang} proposes a self-supervised learning approach for medical image segmentation, utilizing adaptive local prototype networks. It improves feature representations through domain-specific pretraining and integrates anatomical priors for robust prototype construction. Subsequently, Q-Net~\cite{Shen2023} advanced the field by incorporating query-guided prototype refinement, employing anatomical consistency constraints and multi-scale prototype aggregation tailored for medical imaging. More recently, DSPNet~\cite{Tang2024} proposed dynamic support prototype networks, which adapt prototype representations based on query image characteristics through learnable selection and refinement mechanisms. 
While these methods demonstrate sophisticated feature learning capabilities that may capture spatial information, they primarily rely on uniform spatial averaging during prototype extraction, which limits the preservation of boundary-specific information crucial for precise medical segmentation.

\paragraph{Spatial-Aware and Multi-Scale Approaches}
To address the inherent geometric complexities of medical images, sophisticated spatial modeling techniques have progressively emerged. These methods acknowledge that accurate medical segmentation necessitates a deep understanding of spatial relationships and anatomical structure, extending beyond mere feature matching to integrate geometric priors. 
RPT~\cite{RPT} proposed a Region-enhanced Prototypical Transformer for few-shot medical image segmentation, generating regional prototypes from support images and utilizing Bias-alleviated Transformer blocks to manage intra-class diversity. However, RPT exhibits high sensitivity to support set quality, and its multiple regional prototype processing pipelines contribute to considerable computational complexity.
In parallel, recent attention-based approaches enhance feature discrimination and spatial awareness in medical segmentation. CAT-Net~\cite{Lin2023} uses a cross-masked attention Transformer with iterative refinement, bolstering representation capacity for prototypes and features. While promising for long-range dependencies, these methods often demand substantial computational resources and may not sufficiently model anatomical geometric constraints.


\begin{figure*}[t]
  \centering
   \includegraphics[width=\linewidth]{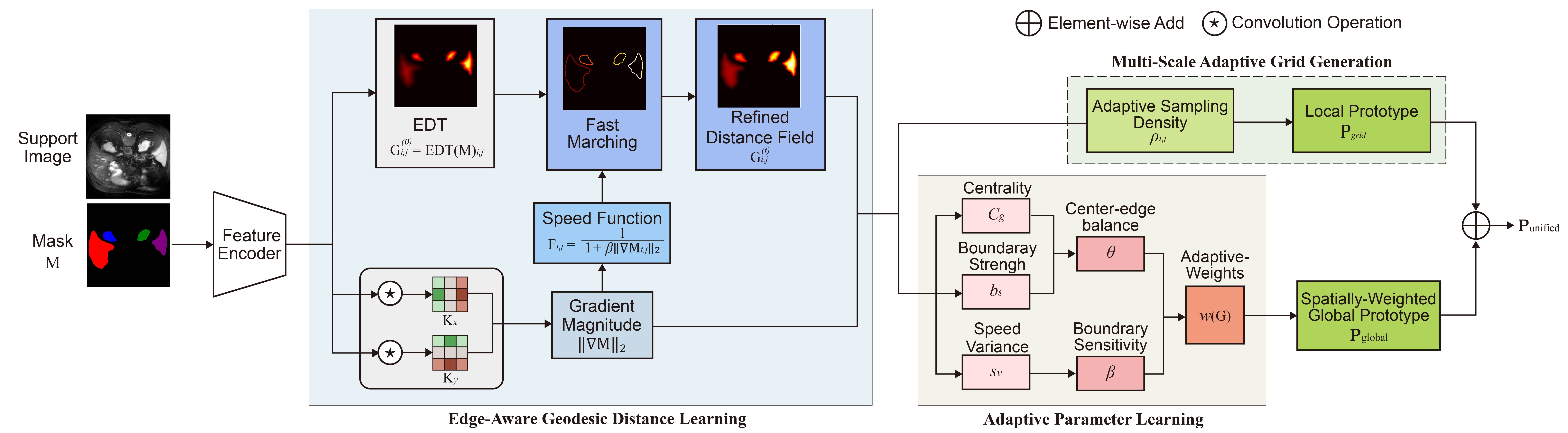}
   \caption{Pipeline of AGENet. Our approach fundamentally transforms prototype extraction by incorporating spatial importance through geodesic distance fields, which respect anatomical boundaries and spatial guidance.}
   \label{fig:pipeline}
\end{figure*}


\section{Method}

Few-shot medical image segmentation aims to segment anatomical structures in query images using only limited support examples. Given a support set $\mathcal{S} = \{(I_i^s, M_i^s)\}_{i=1}^K$ containing $K$ support images with corresponding masks, the goal is to predict the segmentation mask $M^q$ for a query image $I^q$. The fundamental challenge lies in capturing spatial relationships and anatomical structures while generalizing from limited examples. 

This paper introduces AGENet, which enhances prototype-based methods~\cite{Snell17, PANet} by incorporating spatial anatomical information that is not explicitly modeled in uniform averaging approaches, leading to more boundary-aware prototypes for precise organ delineation.
Edge-Aware Geodesic Distance Learning is proposed to transform uniform spatial treatment into anatomically-informed distance fields that respect organ boundaries. It establishes anatomically-aware spatial understanding through: (1) EDT initialization for numerical stability~\cite{Felzenszwalb12}, (2) edge-aware speed functions that slow propagation at detected boundaries, and (3) iterative Fast Marching refinement~\cite{Sethian99} with adaptive parameter learning that automatically adjusts to diverse anatomical structures~\cite{Fedorov12, Litjens17}.
Adaptive Grid Prototype Extraction is proposed to construct spatially-informed prototypes that capture both local and global anatomical patterns. Since single global prototypes cannot capture spatial heterogeneity~\cite{Chen18}, it implements dual prototypes: spatially-weighted global prototypes using geodesic-modulated weighting, and adaptive grid prototypes with boundary-concentrated sampling density.
The components work synergistically: geodesic distance learning provides anatomically-informed spatial fields that guide grid prototype extraction in constructing clean, representative prototypes,  as shown in Figure~\ref{fig:pipeline}. This transforms uniform spatial treatment into anatomically-aware prototype extraction, solving the spatial awareness problem through lightweight geometric modeling without complex overhead.


\subsection{Edge-Aware Geodesic Distance Learning}

\subsubsection{Edge-Aware Computation}

\paragraph{EDT Initialization}
The Euclidean Distance Transform provides robust initialization for medical segmentation masks~\cite{Maurer03,Felzenszwalb12}. 
The theoretical properties of this initialization are established in Definition~\ref{def:edt_edge} (see Appendix~\ref{sec:edt_analysis}).
Given a binary mask $\mathbf{M} \in \{0,1\}^{H \times W}$ where $(i,j)$ denotes pixel coordinates, we initialize the distance field:
\begin{equation}
\mathbf{G}^{(0)}_{i,j} = \text{EDT}(\mathbf{M})_{i,j}
\label{eq:edt_init}
\end{equation}
This initialization ensures numerical stability and provides a baseline distance metric that can be subsequently refined through geometric constraints~\cite{Rosenfeld66,Borgefors86}.

\begin{figure}[H]
  \centering
    \includegraphics[width=\linewidth]{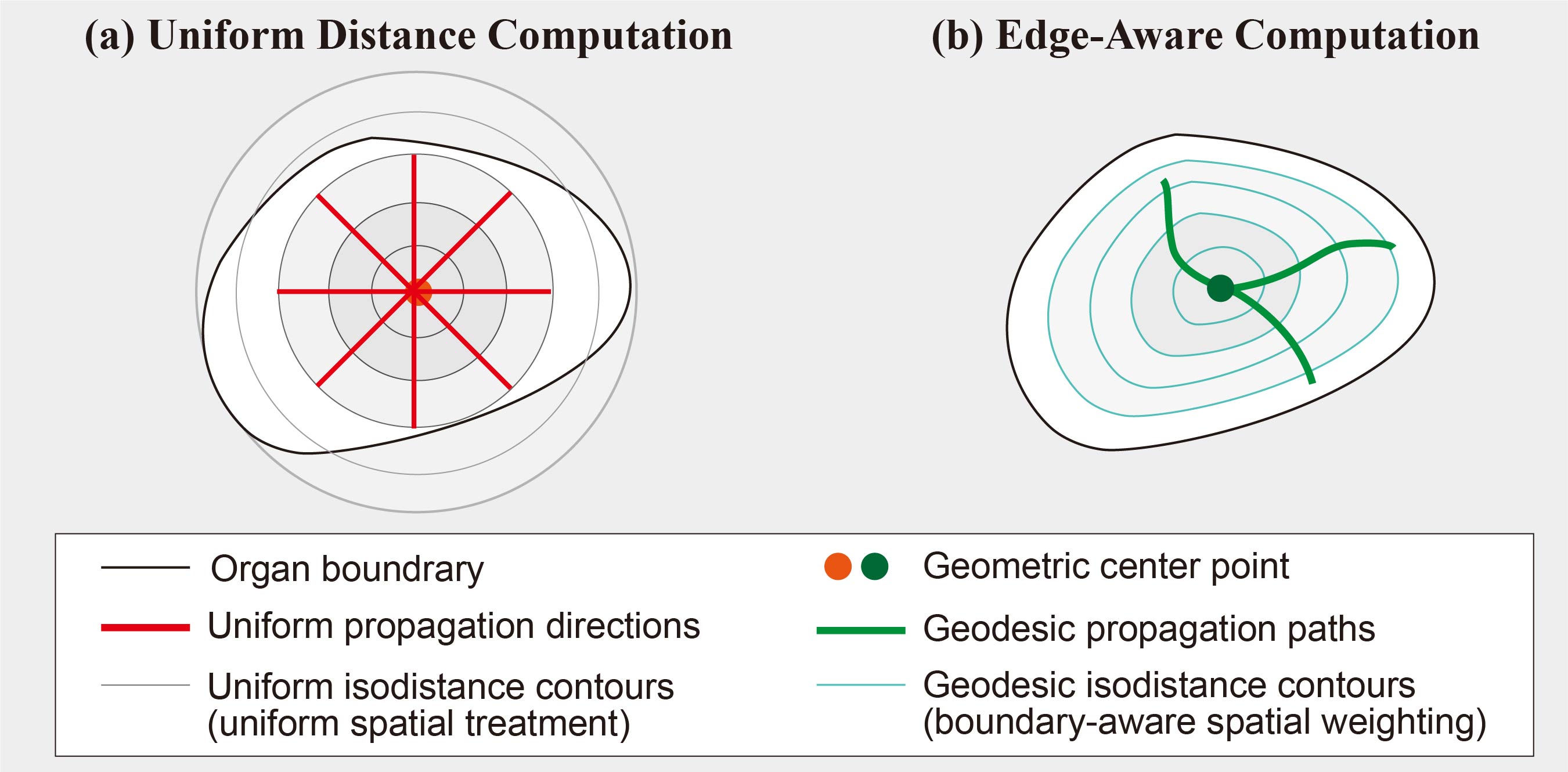}
    \caption{Comparison of distance computation. (a) Uniform distance computation creates regular propagation patterns treating all pixels equally. (b) Edge-aware computation produces boundary-aware spatial weighting that respects anatomical constraints.}
    \label{fig:geo}
\end{figure}

\paragraph{Edge Detection} Medical organs contain internal anatomical structures (vascular networks, tissue interfaces) that create intensity variations in medical images. Traditional prototype extraction treats all organ pixels uniformly, resulting in mixed prototypes that combine organ tissue with transitional boundary features. As shown in Figure~\ref{fig:geo} (a), uniform spatial averaging produces regular distance propagation patterns which not fully capture anatomical constraints. During query segmentation, these mixed prototypes cause similarity computation inaccuracies at organ boundaries, leading to imprecise segmentation. To extract cleaner prototypes that emphasize representative organ tissue, we detect anatomical boundaries using Sobel operators:
\begin{equation}
\|\nabla \mathbf{M}\|_2 = \sqrt{(\mathbf{K}_x * \mathbf{M})^2 + (\mathbf{K}_y * \mathbf{M})^2 + \epsilon}
\label{eq:gradient_magnitude}
\end{equation}
where $\mathbf{K}_x$ and $\mathbf{K}_y$ are Sobel kernels and $\epsilon = 1\text{e-}8$ prevents numerical instability.

\paragraph{Speed Function} 

Following the theoretical framework of level set methods~\cite{Osher88,Malladi95}, the speed function implements the principle that geodesic paths should respect anatomical boundaries. The speed function is formulated as:
\begin{equation}
\mathbf{F}_{i,j} = \frac{1}{1 + \boldsymbol{\beta} \|\nabla \mathbf{M}_{i,j}\|_2}
\label{eq:speed_function}
\end{equation}
where $\beta \in [0.5, 5.0]$ is a learnable parameter controlling boundary sensitivity. The theoretical bounds for this speed function are established in Lemma~\ref{lem:speed_bounds} (see Appendix~\ref{sec:geodesic_analysis}). This formulation is inspired by edge-stopping functions in anisotropic diffusion~\cite{Perona90}, ensuring that geodesic propagation slows near high-gradient regions corresponding to anatomical boundaries (Figure~\ref{fig:geo} (b)).
By reducing speed at anatomical boundaries, this mechanism redistributes spatial weights to de-emphasize transitional regions and amplify homogeneous organ tissue for cleaner prototype extraction.

\subsubsection{Iterative Distance Field Refinement}

The initialized distance field is refined through iterative updates that incorporate edge-aware constraints, following the discrete approximation of the Eikonal equation~\cite{Rouy92,Tsitsiklis95}. This refinement process transforms the detected internal boundaries into spatially-varying distance weights that guide prototype extraction. Starting from the EDT initialization $\mathbf{G}^{(0)}_{i,j} = \text{EDT}(\mathbf{M})_{i,j}$, we compute the speed-modulated update term:
\begin{equation}
\mathbf{U}^{(t)}_{i,j} = \min_{(k,l) \in \mathcal{N}(i,j)} \mathbf{G}^{(t-1)}_{k,l} + \frac{\lambda}{F_{i,j} + \epsilon}
\label{eq:update_term}
\end{equation}
where $\mathcal{N}(i,j)$ represents the 3×3 neighborhood of pixel $(i,j)$, $\lambda = 0.1$ is the step size parameter, and $\epsilon = 10^{-6}$ prevents division by zero. At each iteration $t \geq 1$:
\begin{equation}
\mathbf{G}^{(t)}_{i,j} = \begin{cases} 
\mu \mathbf{G}^{(t-1)}_{i,j} + (1-\mu) \mathbf{U}^{(t)}_{i,j}, & \text{if } \mathbf{M}_{i,j} > \tau \\[0.5em]
\mathbf{G}^{(t-1)}_{i,j}, & \text{otherwise}
\end{cases}
\label{eq:iterative_update}
\end{equation}
where $\mu = 0.7$ balances stability and convergence, $\tau = 0.5$ is the mask threshold, and $T = 3$ Fast Marching iterations balance computational efficiency with convergence quality. Convergence and stability are proven in Theorem~\ref{thm:convergence} and Lemma~\ref{lem:numerical_stability} (see Appendix~\ref{sec:geodesic_analysis}).

\subsubsection{Adaptive Parameter Learning}

Different anatomical structures exhibit distinct spatial characteristics that require tailored processing approaches. Compact organs like kidneys benefit from center-focused weighting, while complex structures like liver demand boundary-sensitive analysis. To address this, we develop an adaptive weighting mechanism that automatically determines the optimal center-boundary balance by analyzing geometric properties of the refined geodesic distance field for each specific image.

We measure geometric centrality $c_g = (\mathbf{G} \odot \mathbf{M}).sum() / (\mathbf{M}.sum() + \epsilon)$ and boundary strength $b_s = \sqrt{(\nabla_x \mathbf{M})^2 + (\nabla_y \mathbf{M})^2}.mean()$ to determine the center-edge balance parameter:
\begin{equation}
\theta = \frac{c_g}{c_g + b_s + \epsilon}
\label{eq:adaptive_theta}
\end{equation}
From the speed function variance $s_v = F.var()$ (see Eq.~\ref{eq:speed_function}), we compute the boundary sensitivity parameter:
\begin{equation}
\beta = \beta_{\text{base}} + \frac{\lambda}{s_v + \epsilon}
\label{eq:adaptive_beta}
\end{equation}
where $\lambda = 4$ is the speed variance scaling factor. These adaptive parameters control the geodesic weighting function that guides prototype extraction:
\begin{equation}
w(\mathbf{G}) = \theta \cdot \tanh(\beta \cdot \mathbf{G}) + (1-\theta) \cdot (1 - \tanh(\beta \cdot \mathbf{G}))
\label{eq:adaptive_weighting}
\end{equation}
There are two key geometric hyperparameters for ablation studies (detailed in Hyperparameter analysis~\ref{Hyperparameter analysis}): Fast Marching iterations $T$ and speed variance scaling factor $\lambda$.

\subsection{Adaptive Grid Prototype Extraction}

\paragraph{Global Prototype Extraction}
We extract spatially-weighted global prototypes using the adaptive weighting function defined in Eq.~\ref{eq:adaptive_weighting}. The global prototype is computed as:
\begin{equation}
\mathbf{p}_{\text{global}} = \frac{\sum_{i,j} \mathbf{M}_{i,j} \cdot w(\mathbf{G}_{i,j}) \cdot \mathbf{F}_{i,j}}{\sum_{i,j} \mathbf{M}_{i,j} \cdot w(\mathbf{G}_{i,j}) + \epsilon}
\label{eq:global_prototype}
\end{equation}
where $\mathbf{F}_{i,j}$ represents the support feature map and $\mathbf{M}_{i,j}$ is the support mask.

\paragraph{Adaptive Grid Generation and Prototype Integration}
Single global prototypes cannot adequately represent spatial heterogeneity of complex medical structures~\cite{Chen18}. We construct an adaptive prototype grid that concentrates sampling density near anatomical boundaries using boundary-aware geodesic weighting. Each grid cell prototype $\mathbf{P}_{g,h}$ aggregates local features within its neighborhood region $\mathcal{N}_{g,h}$:
\begin{equation}
\mathbf{P}_{grid} = \frac{\sum_{(i,j) \in \mathcal{N}_{g,h}} \rho_{i,j} \cdot w_{i,j} \cdot \mathbf{M}_{i,j} \cdot \mathbf{F}_{i,j}}{\sum_{(i,j) \in \mathcal{N}_{g,h}} \rho_{i,j} \cdot w_{i,j} \cdot \mathbf{M}_{i,j} + \epsilon}
\label{eq:adaptive_grid_prototype}
\end{equation}
where the adaptive sampling density is computed as $\rho_{i,j} = 1 + \sigma \cdot \tau \cdot \|\nabla \mathbf{G}_{i,j}\|_2$, with $\sigma \in [1.0, 4.0]$ being the density scale parameter and $\tau \in [0.5, 3.0]$ controlling boundary sensitivity. Both parameters are learned adaptively based on local image characteristics. $\mathcal{N}_{g,h}$ denotes the spatial neighborhood for grid cell $(g,h)$, and $w_{i,j} = w(\mathbf{G}^{(final)}_{i,j})$ applies the adaptive geodesic weighting function from Eq.~\ref{eq:adaptive_weighting}. The bounds for this adaptive density are proven in Appendix Lemma~\ref{lem:adaptive_density} (see Appendix~\ref{sec:prototype_analysis}).

Following hierarchical representation principles~\cite{Vinyals16}, we combine both global and grid prototypes for comprehensive spatial representation:
\begin{equation}
\mathbf{P}_{\text{unified}} = \{\mathbf{P}_{\text{grid}}\} \cup \{\mathbf{p}_{\text{global}}\}
\label{eq:unified_prototypes}
\end{equation}
where $\mathbf{P}_{\text{grid}}$ represents the collection of grid prototypes and $\mathbf{p}_{\text{global}}$ is the global prototype.


\subsection{Loss Function}

To achieve robust few-shot medical image segmentation, we design a comprehensive three-component loss framework that addresses segmentation accuracy, boundary precision, and feature alignment. Further ablation study for loss components can be found in loss function ablation analysis~\ref{Loss Function Ablation}.

\paragraph{Segmentation Loss}

Our base segmentation loss combines cross-entropy and Dice loss with equal weighting to leverage their complementary strengths:
\begin{equation}
\mathcal{L}_{\text{seg}} = \mathcal{L}_{\text{CE}} + \mathcal{L}_{\text{Dice}}
\end{equation}
where the cross-entropy loss is defined as:
\begin{equation}
\mathcal{L}_{\text{CE}} = -\frac{1}{N} \sum_{i=1}^{N} \sum_{c=1}^{C} y_{i,c} \log(\hat{y}_{i,c})
\end{equation}
and the Dice loss is:
\begin{equation}
\mathcal{L}_{\text{Dice}} = 1 - \frac{2\sum_{i=1}^{N} y_{i} \hat{y}_{i} + \epsilon}{\sum_{i=1}^{N} y_{i} + \sum_{i=1}^{N} \hat{y}_{i} + \epsilon}
\end{equation}
Cross-entropy provides pixel-wise classification supervision, while Dice loss directly optimizes the overlap metric commonly used in medical image evaluation.

\paragraph{Edge-Aware Loss}

To enhance boundary accuracy, we introduce an edge-aware loss that emphasizes precise segmentation at organ boundaries using Sobel edge detection:
\begin{equation}
\mathcal{L}_{\text{edge}} = \frac{1}{HW} \sum_{i,j} w_{i,j} \cdot \text{BCE}(E_{\text{pred}}^{i,j}, E_{\text{target}}^{i,j})
\end{equation}
where $E_{\text{pred}}$ and $E_{\text{target}}$ are the predicted and target edge maps respectively, computed using Sobel operators.

\paragraph{Alignment Loss}

To ensure consistency between support and query representations, we employ an alignment loss that enforces bidirectional consistency~\cite{PANet}. This component computes pseudo-labels from query predictions and uses them to re-evaluate support features:
\begin{equation}
\mathcal{L}_{\text{align}} = \frac{1}{NW} \sum_{n=1}^{N} \sum_{w=1}^{W} \mathcal{L}_{\text{CE}}(\text{pred}_{n,w}^{\text{support}}, \text{label}_{n,w}^{\text{pseudo}})
\end{equation}
where $N$ is the number of shots, $W$ is the number of ways, and pseudo-labels are generated from query predictions. This consistency prevents overfitting to support examples and improves generalization to query images. The complete loss function integrates all three components:
\begin{equation}
\mathcal{L}_{\text{total}} = \mathcal{L}_{\text{seg}} +  \mathcal{L}_{\text{edge}} + \mathcal{L}_{\text{align}}
\end{equation}


\section{Experiments}
\label{sec:Experiments}

\subsection{Metrics and Datasets}

\paragraph{Datasets} To evaluate AGENet across diverse medical imaging scenarios, we conducted experiments on nine datasets representing different clinical challenges and data scarcity levels (as shown in Table ~\ref{tab:datasets}). 

\begin{table}[H]
\centering
\caption{Overview of datasets used for few-shot medical image segmentation evaluation.}
\fontsize{8pt}{10pt}\selectfont
\label{tab:datasets}
\begin{tabular}{p{3cm}p{2.5cm}p{1.5cm}}
\hline
Dataset & Modality/Target & Total \\
\hline
ABD-MRI~\cite{ABD-MRI} & MRI Abdomen & 20 scans \\
ABD-CT~\cite{ABD-CT} & CT Abdomen & 30 scans \\
ACDC~\cite{ACDC} & MRI Heart & 150 scans \\
PH2 (Atypical Nevus)~\cite{PH2} & Dermoscopy Skin & 80 images \\
PH2 (Melanomas)~\cite{PH2} & Dermoscopy Skin & 40 images \\
CVC-300~\cite{CVC300} & Endoscopy Colon & 60 images \\
CHASE-DB1~\cite{CHASE-DB1} & Fundus Vessels & 28 images \\
Drishti-GS~\cite{Drishti-GS} & Fundus Glaucoma & 70 images \\
DCA1~\cite{DCA1} & X-ray Coronary & 134 images \\
\hline
\end{tabular}
\end{table}

\begin{table*}[h]
\centering
\caption{Cross-domain segmentation performance comparison across diverse medical imaging datasets. Results show DICE coefficient (\%) and 95th percentile Hausdorff Distance (HD95) in mm. \textbf{Bold} indicates best performance among all methods.}
\label{tab:cross-domain-performance}
\fontsize{5pt}{6.5pt}\selectfont
\resizebox{\textwidth}{!}{%
\begin{tabular}{p{0.8cm}|l|ccccccccc|c}
\hline
\rotatebox{90}{Metric} & \rotatebox{90}{Method} & \rotatebox{90}{\begin{tabular}{@{}c@{}}PH2(Atypical\\Nevus)\end{tabular}} & \rotatebox{90}{\begin{tabular}{@{}c@{}}PH2\\(Melanomas)\end{tabular}} & \rotatebox{90}{CVC-300} & \rotatebox{90}{CHASE-DB1} & \rotatebox{90}{Drishti-GS} & \rotatebox{90}{DCA1} & \rotatebox{90}{ABD-MRI} & \rotatebox{90}{ABD-CT} & \rotatebox{90}{ACDC} & \rotatebox{90}{Avg} \\
\hline
\multirow{9}{*}{\begin{tabular}{c}DICE \\ (\%)\end{tabular}} 
 & PANet~\cite{PANet} & 82.81 & 79.96 & 81.21 & 38.74 & 84.98 & 43.17 & 36.61 & 55.00 & 39.00 & 60.16 \\
 & SE-Net~\cite{Roy2020} & 78.23 & 75.37 & 76.11 & 32.59 & 81.87 & 38.32 & 42.73 & 28.18 & 30.20 & 53.73 \\
 & SSL-ALPNet~\cite{Ouyang} & 87.44 & 84.77 & 85.92 & 46.89 & 89.32 & 52.66 & 72.96 & 70.58 & 61.20 & 72.42 \\
 & Q-Net~\cite{Shen2023} & 88.13 & 85.46 & 86.67 & 47.35 & 89.81 & 53.92 & 75.20 & 61.95 & 58.50 & 71.89 \\
 & CAT-Net~\cite{Lin2023} & 87.76 & 84.99 & 86.12 & 46.93 & 89.55 & 53.24 & 72.60 & 61.88 & 57.40 & 71.16 \\
 & DSPNet~\cite{Tang2024} & 90.97 & 87.16 & 88.31 & 50.42 & 91.21 & 56.82 & 79.29 & 72.79 & 64.80 & 75.75 \\
 & RPT~\cite{RPT} & 90.36 & 88.68 & 89.82 & 52.74 & 92.81 & 55.12 & 82.44 & 77.83 & 68.40 & 77.24 \\
\rowcolor{gray!20} 
 & \textbf{Ours-1-shot} & 94.29 & 90.99 & 91.82 & 51.81 & 94.08 & 61.47 & 82.60 & 77.60 & 69.70 & 79.56 \\
\rowcolor{gray!20} 
 & \textbf{Ours-5-shot} & \textbf{94.81} & \textbf{92.21} & \textbf{93.62} & \textbf{54.21} & \textbf{97.39} & \textbf{62.93} & \textbf{85.21} & \textbf{80.78} & \textbf{71.92} & \textbf{81.67} \\
\hline
\multirow{9}{*}{\begin{tabular}{c}HD95 \\ (mm)\end{tabular}} 
 & PANet~\cite{PANet} & 13.15 & 19.28 & 15.64 & 29.93 & 5.17 & 26.35 & 27.69 & 35.39 & 24.16 & 21.86 \\
 & SE-Net~\cite{Roy2020} & 15.83 & 23.47 & 18.25 & 35.72 & 7.34 & 31.86 & 31.85 & 33.37 & 33.04 & 25.64 \\
 & SSL-ALPNet~\cite{Ouyang} & 10.94 & 16.73 & 11.82 & 24.67 & 3.74 & 21.25 & 18.04 & 19.08 & 17.98 & 16.03 \\
 & Q-Net~\cite{Shen2023} & 10.34 & 15.89 & 11.45 & 24.23 & 3.62 & 20.94 & 16.69 & 22.69 & 20.02 & 16.21 \\
 & CAT-Net~\cite{Lin2023} & 10.72 & 16.34 & 12.18 & 22.84 & 3.85 & 19.46 & 17.74 & 23.07 & 19.66 & 16.21 \\
 & DSPNet~\cite{Tang2024} & 9.17 & 14.82 & 10.26 & 21.43 & 3.14 & 18.34 & 25.32 & 18.12 & 16.55 & 15.24 \\
 & RPT~\cite{RPT} & 8.86 & 13.95 & 10.94 & 22.17 & 3.35 & 18.84 & 12.72 & 16.06 & 17.96 & 13.87 \\
\rowcolor{gray!20} 
 & \textbf{Ours-1-shot} & 8.21 & 10.01 & 10.01 & 16.26 & 2.77 & 12.32 & 14.06 & 14.83 & 11.98 & 11.16 \\
\rowcolor{gray!20} 
 & \textbf{Ours-5-shot} & \textbf{7.64} & \textbf{4.68} & \textbf{4.68} & \textbf{13.08} & \textbf{1.61} & \textbf{8.02} & \textbf{12.60} & \textbf{12.92} & \textbf{10.29} & \textbf{8.39} \\
\hline
\end{tabular}%
}
\end{table*}

\begin{figure}[t]
  \centering
   \includegraphics[width=\linewidth]{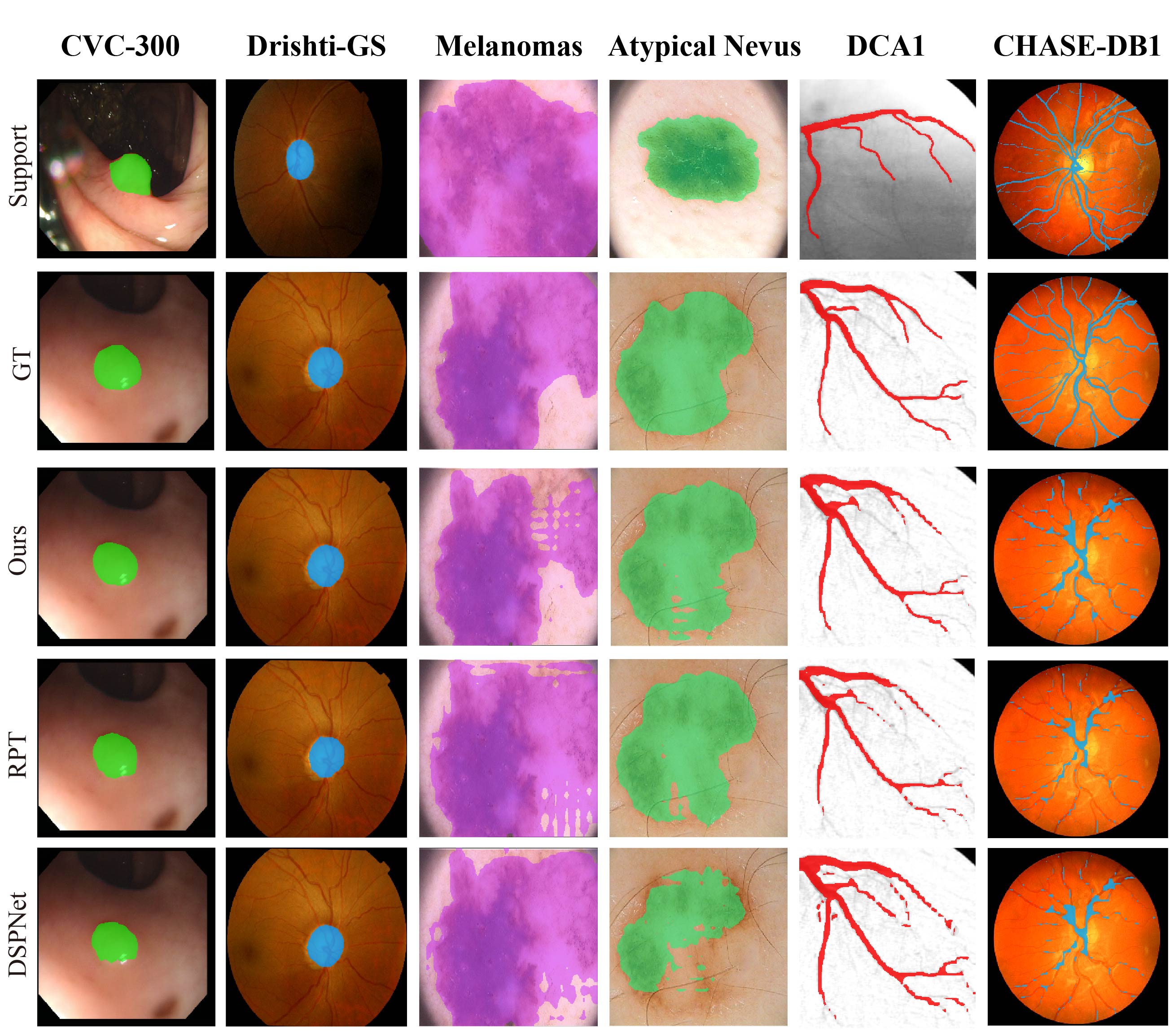}
   \caption{Visualization from left to right on CVC-300, Drishti-GS, PH2(Melanomas), PH2(Atypical Nevus), DCA1 and CHASE-DB1 datasets. Top row to bottom: Support, Ground-truth. Ours prediction, RPT's prediction and DSPNet's prediction. }
   \label{fig:organ-v1}
\end{figure}

Our evaluation spans multiple clinical specialties including cardiac imaging with ACDC~\cite{ACDC} (150 MRI scans) and abdominal imaging with ABD-MRI~\cite{ABD-MRI} (20 MRI scans) and ABD-CT~\cite{ABD-CT} (30 CT scans). We evaluate on applications with complex morphological structures where expert annotations are extremely expensive, including interventional cardiology with DCA1~\cite{DCA1} (134 coronary artery X-ray images) and retinal vessel segmentation with CHASE-DB1~\cite{CHASE-DB1} (28 fundus images). We assess performance on small-scale datasets including CVC-300~\cite{CVC300} endoscopy (60 colon images) and Drishti-GS~\cite{Drishti-GS} (70 glaucoma images). We evaluate on rare dermatological conditions using PH2~\cite{PH2} (40 melanoma and 80 atypical nevus images). Further detailed description can be found in Appendix~\ref{appendix:Datasets Description}.

\paragraph{Evaluation Metrics} We employ comprehensive evaluation metrics following standard medical image segmentation protocols~\cite{Taha2015}. The \textbf{DICE Similarity Coefficient (DSC)}~\cite{DICE1945} measures region overlap between predicted and ground truth segmentations. 
For boundary quality, we use the \textbf{95th percentile Hausdorff Distance (HD95)}~\cite{Huttenlocher1993}, which measures maximum surface distance errors while being robust to outliers. These metrics provide evaluation of both region-based accuracy and boundary precision.

\paragraph{Implementation Details}

We implement our network with PyTorch, using pre-trained ResNet-101 based DeepLab encoder initialized on MS-COCO dataset. We scale input images to 3×256×256, obtaining feature maps of 256×32×32 after encoding. The adaptive prototype grid operates at 8×8 resolution. We use SGD optimizer with initial learning rate 0.005 and multi-step scheduler reducing by 0.95. We set batch size to 1 and train for 100,000 iterations. The model trains on NVIDIA RTX 4090 GPU for around 4 hours per fold.
For all experiments, we follow the protocol from~\cite{Ouyang}. All baseline methods are evaluated in the 1-shot setting, and we report both 1-shot and 5-shot results for our method.

\begin{table*}[t]
\centering
\caption{DICE coefficient (\%) performance comparison on ABD-MRI, ABD-CT, and ACDC datasets. \textbf{Bold} indicates best performance.}
\label{tab:dice_performance}
\fontsize{8pt}{10pt}\selectfont
\resizebox{\textwidth}{!}{%
\begin{tabular}{l|ccccc|ccccc|cccc}
\hline
\multirow{2}{*}[-0.5ex]{Method} & \multicolumn{5}{c|}{ABD-MRI} & \multicolumn{5}{c|}{ABD-CT} & \multicolumn{4}{c}{ACDC} \\
 & Liver & RK & LK & Spleen & Avg & Liver & RK & LK & Spleen & Avg & RV & MYO & LV & Avg \\
\hline

PANet~\cite{PANet} & 48.02 & 32.15 & 36.40 & 29.88 & 36.61 & 59.12 & 51.80 & 54.67 & 54.41 & 55.00 & 35.80 & 35.40 & 45.90 & 39.03 \\
SE-Net~\cite{Roy2020} & 30.11 & 48.20 & 44.85 & 45.76 & 42.73 & 34.05 & 13.20 & 25.38 & 42.10 & 28.18 & 27.70 & 27.40 & 35.50 & 30.20 \\
SSL-ALPNet~\cite{Ouyang} & 69.55 & 78.80 & 80.30 & 63.20 & 72.96 & 66.80 & 71.90 & 74.60 & 69.01 & 70.58 & 56.10 & 55.50 & 72.00 & 61.20 \\
Q-Net~\cite{Shen2023} & 72.20 & 83.10 & 76.80 & 70.70 & 75.20 & 67.10 & 54.20 & 68.30 & 58.20 & 61.95 & 53.60 & 53.10 & 68.80 & 58.50 \\
CAT-Net~\cite{Lin2023} & 71.00 & 77.40 & 74.60 & 67.40 & 72.60 & 65.80 & 48.40 & 68.10 & 65.20 & 61.88 & 52.60 & 52.10 & 67.50 & 57.40 \\
DSPNet~\cite{Tang2024} & 74.33 & 87.88 & 85.62 & 69.31 & 79.29 & 69.32 & 74.54 & 78.01 & 69.31 & 72.79 & 59.40 & 58.80 & 76.20 & 64.80 \\
RPT~\cite{RPT} & \textbf{82.86} & 89.92 & 80.72 & 76.37 & 82.44 & \textbf{82.77} & 77.05 & 79.13 & 72.58 & 77.83 & 62.70 & 66.00 & 80.40 & 69.70 \\
\rowcolor{gray!20} \textbf{Ours-1-shot} & 77.97 & 91.45 & 87.04 & 73.94 & 82.60 & 78.81 & 79.02 & 81.07 & 71.50 & 77.60 & 63.51 & 63.05 & 81.59 & 69.38 \\
\rowcolor{gray!20} \textbf{Ours-5-shot} & 79.52 & \textbf{91.52} & \textbf{87.79} & \textbf{81.99} & \textbf{85.21} & 81.34 & \textbf{82.17} & \textbf{84.92} & \textbf{74.68} & \textbf{80.78} & \textbf{66.06} & \textbf{67.06} & \textbf{82.63} & \textbf{71.92} \\
\hline
\end{tabular}%
}
\end{table*}

\begin{figure*}[t]
  \centering
   \includegraphics[width=\textwidth]{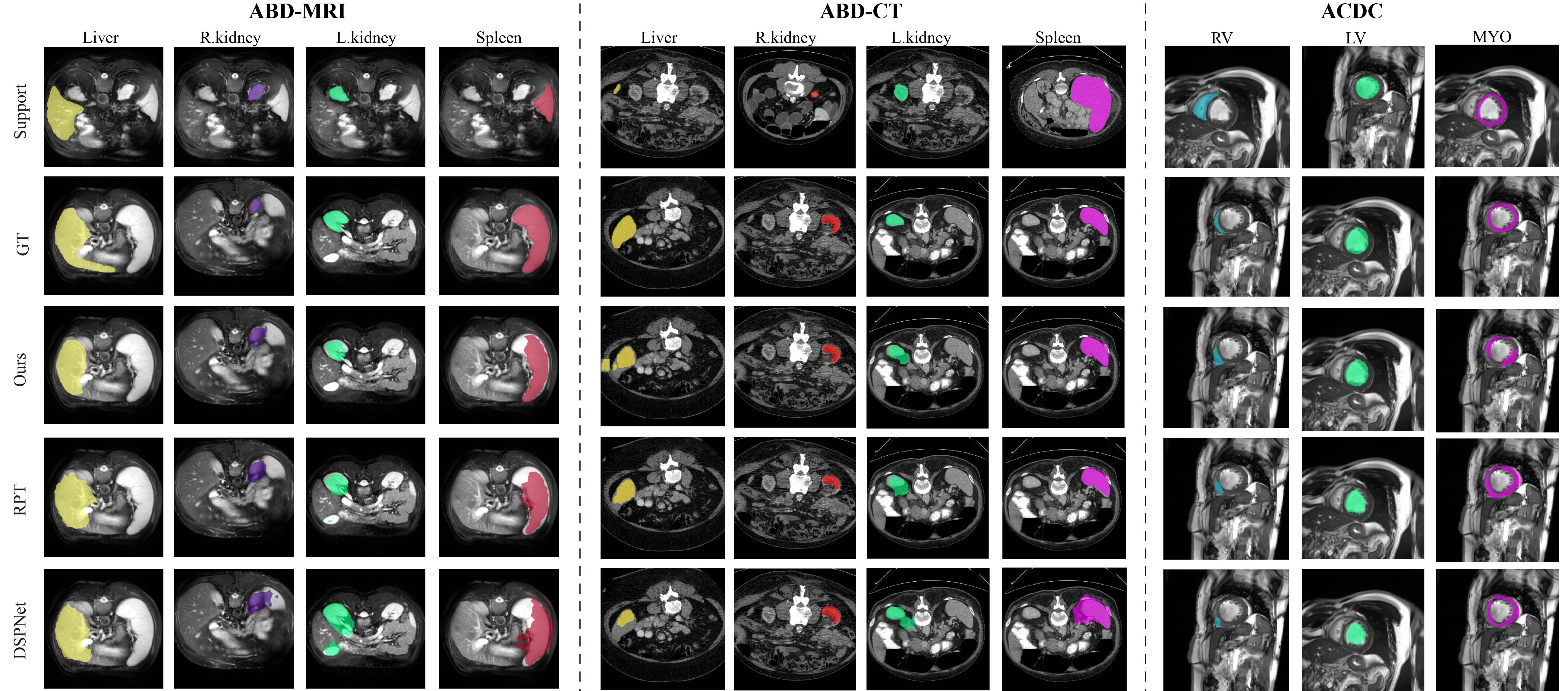}
   \caption{Visualization from left to right on ABD-MRI, ABD-CT and ACDC datasets with all organs. Top row to bottom: Support, Ground-truth. Ours prediction, RPT's prediction and DSPNet's prediction.}
   \label{fig:3d-visual}
\end{figure*}

\subsection{Comparison with State-of-the-Art Methods}

We compare against state-of-the-art prototype-based methods including PANet~\cite{PANet}, SE-Net~\cite{Roy2020}, SSL-ALPNet~\cite{Ouyang}, Q-Net~\cite{Shen2023}, CAT-Net~\cite{Lin2023}, DSPNet~\cite{Tang2024} and RPT~\cite{RPT}.

\paragraph{Performance Comparison}  Table~\ref{tab:cross-domain-performance} presents cross-domain evaluation results on nine different medical imaging datasets, demonstrating our method's superior generalization capabilities. Notably, our method maintains consistent improvements on challenging datasets like CHASE-DB1 (vessel segmentation) and DCA1 (coronary arteries) with complex vascular structures. Figure~\ref{fig:organ-v1} and Appendix~\ref{Qualitative results} show qualitative visualization results on six representative datasets, illustrating the visual superiority of our predictions over competing methods. 5-shot approach achieves the highest average DICE coefficient of 81.67\% and the lowest average HD95 distance of 8.39mm, while our 1-shot method also demonstrates strong performance with 79.56\% DICE and 11.16mm HD95, both outperforming existing methods including RPT (77.24\% DICE, 13.87mm HD95) and DSPNet (75.75\% DICE, 15.24mm HD95).

Table~\ref{tab:dice_performance} and Figure~\ref{fig:3d-visual} provide detailed organ-specific performance analysis on ABD-MRI, ABD-CT, and ACDC datasets for both our 1-shot and 5-shot configurations. Our 5-shot method consistently achieves superior performance across most anatomical structures, with particularly notable improvements in kidney segmentation (91.52\% DICE for right kidney in ABD-MRI) and cardiac structures (82.63\% DICE for left ventricle in ACDC). Our 1-shot approach also demonstrates competitive results with average scores of 82.60\%, 77.60\%, and 69.38\% for ABD-MRI, ABD-CT, and ACDC respectively. While RPT shows competitive performance on liver segmentation in ABD-MRI (82.86\% vs our 79.52\% for 5-shot), our method demonstrates more balanced and consistent performance across all organs and datasets, with our 5-shot configuration achieving the highest average scores across all three datasets.

\paragraph{Computational complexity comparison}
\label{Computational complexity comparison}
Our proposed method demonstrates superior computational efficiency compared to existing approaches, as shown in Table~\ref{tab:complexity}. While RPT~\cite{RPT} requires the largest memory footprint (489.21 MB) and computational cost (132.9 GFLOPs) with 58.3M parameters, our method achieves comparable performance with significantly reduced resource consumption. Specifically, our approach uses 46.5\% less memory (261.88 MB) and 32.5\% fewer FLOPs (89.7 G) than RPT, while maintaining competitive parameter efficiency (58.2 M). Although our inference time (11.17 ms) is higher than the highly optimized PANet (3.04 ms), it remains substantially faster than both RPT (30.84 ms) and DSPNet (27.07 ms), and is competitive with other methods like CAT-Net (9.15 ms), striking an optimal balance between computational efficiency and practical deployment requirements.

\begin{table}[H]
\centering
\caption{Computational complexity comparison of methods. }
\label{tab:complexity}
\resizebox{\columnwidth}{!}{%
\fontsize{8pt}{9.5pt}\selectfont
\begin{tabular}{lcccc}
\hline
Method & \begin{tabular}{@{}c@{}}Memory \\ (MB)\end{tabular} & \begin{tabular}{@{}c@{}}FLOPs \\ (G))\end{tabular} & \begin{tabular}{@{}c@{}}Params \\ (M)\end{tabular} & \begin{tabular}{@{}c@{}}Inference\\Time (ms)\end{tabular} \\
\hline
PANet~\cite{PANet} & 186.51 & 69.1 & 14.7 & 3.04 \\
SSL-ALPNet~\cite{Ouyang} & 289.11 & 134.7 & 59.2 & 7.286 \\
Q-Net~\cite{Shen2023} & 235.66 & 132.7 & 46.1 & 7.03 \\
CAT-Net~\cite{Lin2023} & 267.45 & 98.4 & 52.8 & 12.15 \\
DSPNet~\cite{Tang2024} & 298.11 & 91.3 & 59.3 & 27.07 \\
RPT~\cite{RPT} & 489.21 & 132.9 & 58.3 & 30.84 \\
\rowcolor{gray!20} Ours & 261.88 & 89.7 & 58.2 & 11.17 \\
\hline
\end{tabular}%
}
\end{table}


\subsection{Ablation Study}

\paragraph{Module Ablation Analysis}
The module ablation study (Table~\ref{tab:module_ablation}) systematically evaluates the contribution of each proposed component in AGENet. The results demonstrate that geodesic distance learning (GDL) constitutes the most critical component, with its removal causing substantial performance degradation (10.77 DICE points reduction in 1-shot scenarios, from 79.56 to 68.79). The adaptive grid prototypes (AGP) component shows significant impact on boundary precision, with removal leading to notable performance drops (5.89 DICE points in 1-shot and 5.76 points in 5-shot settings). Each remaining component (EAC, IFM, and APL) contributes meaningfully to the overall framework performance. All components together achieve 10.77 and 9.24 DICE improvements over baseline in 1-shot and 5-shot settings, with HD95 improvements of 11.51 and 10.89 points, validating the multi-component architecture.

\begin{table}[H]
\centering
\caption{Ablation study showing the contribution of each component. \textbf{Bold} indicates best performance. DICE and HD95 values are averaged across all datasets. GDL: geodesic distance learning; EAC: edge-aware computation; IFM: iterative fast marching; AGP: adaptive grid prototypes; APL: adaptive parameters.}
\label{tab:module_ablation}
\resizebox{\columnwidth}{!}{%
\fontsize{12pt}{13.5pt}\selectfont
\begin{tabular}{ccccc|cc|cc}
\hline
\multicolumn{5}{c|}{Components} & \multicolumn{2}{c|}{1-shot} & \multicolumn{2}{c}{5-shot} \\
\hline
GDL & EAC & IFM & AGP & APL & DICE & HD95 & DICE & HD95 \\
\hline
\checkmark & \checkmark & \checkmark & \checkmark & \checkmark & \textbf{79.56} & \textbf{11.16} & \textbf{81.67} & \textbf{8.39} \\
\checkmark & \checkmark & \checkmark & \checkmark & $\times$ & 78.23 & 12.45 & 80.31 & 9.67 \\
\checkmark & \checkmark & $\times$ & \checkmark & $\times$ & 76.42 & 15.27 & 78.65 & 12.13 \\
\checkmark & $\times$ & $\times$ & \checkmark & $\times$ & 75.85 & 18.94 & 77.91 & 16.52 \\
\checkmark & \checkmark & \checkmark & $\times$ & $\times$ & 73.67 & 16.82 & 75.91 & 14.65 \\
\checkmark & $\times$ & $\times$ & $\times$ & $\times$ & 71.45 & 19.27 & 74.12 & 16.94 \\
$\times$ & $\times$ & $\times$ & $\times$ & $\times$ & 68.79 & 22.67 & 72.43 & 19.28 \\
\hline
\end{tabular}%
}
\end{table}

\paragraph{Geometric Hyperparameter analysis}
\label{Hyperparameter analysis}

As shown in Table ~\ref{tab:ablation_parameters}, our optimal configuration $T=3$ iterations and $\lambda=4$ speed variance scaling directly reflects the geometric convergence properties of anatomical structures, achieving 79.56\% (1-shot) and 81.67\% (5-shot) DICE scores as predicted by our theoretical bounds.
$T=3$ represents the geometric convergence point where geodesic distance fields capture anatomical boundaries without over-smoothing. Fewer iterations ($T=1$, 75.82\% DICE) provide insufficient boundary refinement, while excessive iterations ($T=5$, 71.85\% DICE) cause geometric instability through over-correction, violating our convergence guarantees from Theorem B.3.
 $\lambda=4$ optimally balances edge-stopping properties with numerical stability according to our speed function bounds. Parameter deviations systematically degrade the geometric constraints essential for accurate prototype extraction.

\begin{table}[H]
\centering
\caption{Ablation study on iterations $T$ and speed variance scale $\lambda$ parameters. \textbf{Bold} indicates best performance.}
\label{tab:ablation_parameters}
\resizebox{\columnwidth}{!}{%
\fontsize{4.5pt}{5.5pt}\selectfont
\begin{tabular}{cc|cc|cc}
\hline
\multicolumn{2}{c|}{Parameters} & \multicolumn{2}{c|}{1-shot} & \multicolumn{2}{c}{5-shot} \\
\hline
$T$ & $\lambda$ & DICE & HD95 & DICE & HD95 \\
\hline
1 & 2 & 75.82 & 18.96 & 77.61 & 16.21 \\
2 & 3 & 77.20 & 14.75 & 79.85 & 12.10 \\
3 & 4 & \textbf{79.56} & \textbf{11.16} & \textbf{81.67} & \textbf{8.39} \\
4 & 5 & 74.20 & 15.25 & 79.15 & 12.19 \\
5 & 6 & 71.85 & 19.40 & 76.20 & 16.80 \\
\hline
\end{tabular}%
}
\end{table}

\paragraph{Loss Function Ablation Analysis}
\label{Loss Function Ablation}

Table~\ref{tab:ablation_loss_components} demonstrates the contribution of each loss component across 1-shot and 5-shot settings. The baseline segmentation loss $\mathcal{L}_{seg}$ achieves 72.30\% DICE (1-shot) and 75.80\% DICE (5-shot). Adding alignment loss $\mathcal{L}_{align}$ improves performance to 76.15\% and 79.40\% respectively, while edge loss $\mathcal{L}_{edge}$ yields 74.80\% and 77.60\%. The full combination of all three components achieves optimal results with 79.56\% DICE and 11.16mm HD95 for 1-shot, and 81.67\% DICE with 8.39mm HD95 for 5-shot, confirming the synergistic effect of our proposed loss design.

\begin{table}[H]
\centering
\caption{Ablation study on loss function. DICE and HD95 are averaged across all datasets. \textbf{Bold} indicates best performance.}
\label{tab:ablation_loss_components}
\resizebox{\columnwidth}{!}{%
\fontsize{8pt}{9.5pt}\selectfont
\begin{tabular}{ccc|cc|cc}
\hline
\multicolumn{3}{c|}{Loss Function} & \multicolumn{2}{c|}{1-shot} & \multicolumn{2}{c}{5-shot} \\
\hline
$\mathcal{L}_{seg}$ & $\mathcal{L}_{align}$ & $\mathcal{L}_{edge}$ & DICE & HD95 & DICE & HD95 \\
\hline
\checkmark & $\times$ & $\times$ & 72.30 & 15.45 & 75.80 & 11.20 \\
\checkmark & \checkmark & $\times$ & 76.15 & 12.95 & 79.40 & 9.25 \\
\checkmark & $\times$ & \checkmark & 74.80 & 14.35 & 77.60 & 10.40 \\
\checkmark & \checkmark & \checkmark & \textbf{79.56} & \textbf{11.16} & \textbf{81.67} & \textbf{8.39} \\
\hline
\end{tabular}%
}
\end{table}


\section{Conclusion}
\label{sec:Conclusion}

We introduce AGENet, a novel edge-aware geodesic distance learning framework that transforms few-shot medical image segmentation through anatomically-informed spatial modeling.
Experimental validation across nine diverse datasets demonstrates substantial improvements over state-of-the-art approaches, achieving 79.56\% and 81.67\% mean DICE scores for 1-shot and 5-shot scenarios respectively. Ablation studies confirm the synergistic contributions of our geodesic distance learning, adaptive prototype extraction, and parameter learning components. Through lightweight geometric modeling, our approach enhances clinical utility by adapting to diverse anatomical structures without complex architectural modifications.

\section*{Acknowledgments}
The author would like to thank Dr. Martyn Harris for his valuable guidance and support for this research.

{
    \small
    \bibliographystyle{ieeenat_fullname}
    \bibliography{main}
}

\clearpage
\appendix
\onecolumn

\section{Datasets Description}
\label{appendix:Datasets Description}

\paragraph{ACDC (2018)}~\cite{ACDC}: Automated Cardiac Diagnosis Challenge dataset containing 150 cardiac MRI scans from different patients with multi-structure segmentation annotations for left ventricle, right ventricle, and myocardium. Acquired over 6 years at University Hospital of Dijon using 1.5T and 3.0T Siemens scanners with SSFP sequences, spatial resolution 1.37-1.68 mm²/pixel, slice thickness 5-8mm, and 28-40 cardiac cycle frames. Dataset encompasses five evenly distributed groups (30 patients each): healthy subjects and four cardiac pathologies including dilated cardiomyopathy, hypertrophic cardiomyopathy, myocardial infarction, and abnormal right ventricle, representing comprehensive cardiac imaging scenarios for automated diagnosis assistance.

\paragraph{ABD-MRI (2021)}~\cite{ABD-MRI}: CHAOS Challenge combined healthy abdominal organ segmentation dataset comprising 40 T2-SPIR weighted MRI scans (20 training cases) with annotations for liver, spleen, left kidney, and right kidney. Acquired using 1.5T Philips MRI producing 12-bit DICOM images at 256×256 resolution with variable slice thickness 5.5-9mm (average 7.84mm), in-plane spacing 1.36-1.89mm (average 1.61mm), and 26-50 slices per volume. Dataset emphasizes cross-modality segmentation challenges and healthy organ delineation from potential liver donors, supporting development of automated abdominal imaging analysis for clinical assessment and surgical planning.

\paragraph{ABD-CT (2015)}~\cite{ABD-CT}: MICCAI Multi-Atlas Labeling Beyond the Cranial Vault dataset containing 50 abdominal CT scans from colorectal cancer chemotherapy trials and ventral hernia studies with multi-organ segmentation masks. Acquired during portal venous contrast phase with variable volumes (512×512×85-198), field of view 280-650mm³, in-plane resolution 0.54-0.98mm², and slice thickness 2.5-5.0mm. Dataset includes clinical cases from patients with metastatic liver cancer or postoperative abdominal wall hernia, providing comprehensive coverage of abdominal imaging challenges encountered in routine clinical practice and radiological interpretation.

\paragraph{PH2 (2013)}~\cite{PH2}: Dermoscopic image database for skin lesion research containing 200 high-resolution (768×560 pixels) 8-bit RGB images acquired at Hospital Pedro Hispano, Portugal using Tuebinger Mole Analyzer with 20× magnification under standardized conditions. Dataset includes 80 atypical nevus, 40 melanoma cases, and 80 common nevi with expert dermatologist annotations including lesion segmentation, clinical assessment, and dermoscopic criteria evaluation (colors, pigment network, dots/globules, streaks). Supporting automated melanoma detection and skin cancer screening applications in dermatological practice with comprehensive clinical and histological diagnosis metadata.

\paragraph{CVC-300 (2017)}~\cite{CVC300}: Endoscopic scene segmentation benchmark containing 60 colonoscopy images with polyp annotations extracted from colonoscopy video sequences. Dataset emphasizes endoluminal scene understanding including polyp detection, lumen segmentation, and specular highlight identification, essential for computer-aided diagnosis during colonoscopy procedures.

\paragraph{CHASE-DB1 (2012)}~\cite{CHASE-DB1}: Retinal vessel segmentation database comprising 28 fundus photography images with manual vessel annotations from two independent human experts. Dataset covers diverse retinal imaging conditions and vessel morphologies, supporting automated diabetic retinopathy screening and cardiovascular risk assessment through retinal vessel analysis.

\paragraph{Drishti-GS (2014)}~\cite{Drishti-GS}: Glaucoma screening dataset containing 70 fundus images with optic nerve head and optic cup segmentation annotations. Dataset emphasizes glaucoma detection challenges including varying optic disc appearances, cup-to-disc ratios, and pathological conditions, enabling automated glaucoma screening and early detection systems.

\paragraph{DCA1 (2019)}~\cite{DCA1}: Coronary artery segmentation dataset comprising 134 X-ray angiography images with vessel annotations. Dataset addresses interventional cardiology challenges including complex vessel morphologies, overlapping structures, and contrast variations typical in coronary angiographic procedures, supporting automated vessel analysis for cardiac intervention planning.

\section{Math Proof}
\subsection{EDT Initialization Analysis}
\label{sec:edt_analysis}

We provide the mathematical foundation for our EDT-based initialization scheme, which ensures robust starting conditions for the subsequent geodesic refinement process.

\begin{definition}[EDT Edge Detection]\label{def:edt_edge}
For a binary mask $\mathbf{M} \in \{0,1\}^{H \times W}$, the EDT initialization incorporates edge detection via Sobel operators. The gradient magnitude is computed as:
\begin{equation}\label{eq:gradient_magnitude}
\|\nabla \mathbf{M}_{i,j}\|_2 = \sqrt{(\mathbf{K}_x * \mathbf{M})_{i,j}^2 + (\mathbf{K}_y * \mathbf{M})_{i,j}^2 + \epsilon}
\end{equation}
where $\mathbf{K}_x = \begin{bmatrix} -1 & 0 & 1 \\ -2 & 0 & 2 \\ -1 & 0 & 1 \end{bmatrix}$ and $\mathbf{K}_y = \begin{bmatrix} -1 & -2 & -1 \\ 0 & 0 & 0 \\ 1 & 2 & 1 \end{bmatrix}$ are Sobel kernels.

The EDT initialization $\mathbf{G}^{(0)}_{i,j} = \text{EDT}(\mathbf{M})_{i,j}$ satisfies:
\begin{enumerate}[label=(\roman*)]
    \item \textbf{Boundary condition:} $\mathbf{G}^{(0)}_{i,j} = 0$ for all $(i,j) \in \partial\mathbf{M}$;
    \item \textbf{Monotonicity:} $\mathbf{G}^{(0)}_{i,j} \leq \mathbf{G}^{(0)}_{k,l}$ whenever $d((i,j), \partial\mathbf{M}) \leq d((k,l), \partial\mathbf{M})$;
    \item \textbf{Lipschitz continuity:} $|\mathbf{G}^{(0)}_{i,j} - \mathbf{G}^{(0)}_{k,l}| \leq \|(i,j) - (k,l)\|_2$.
\end{enumerate}
\end{definition}

\begin{lemma}[Speed Function Bounds]\label{lem:speed_bounds}
The speed function $\mathbf{F}_{i,j} = \frac{1}{1 + \boldsymbol{\beta} \|\nabla \mathbf{M}_{i,j}\|_2}$ satisfies:
\begin{align}\label{eq:speed_function_bounds}
\frac{1}{1 + 5\|\nabla \mathbf{M}_{i,j}\|_2} \leq \mathbf{F}_{i,j} \leq \frac{1}{1 + 0.5\|\nabla \mathbf{M}_{i,j}\|_2}
\end{align}
for $\boldsymbol{\beta} \in [0.5, 5.0]$.
\end{lemma}

\subsection{Geodesic Refinement Analysis}
\label{sec:geodesic_analysis}

This section establishes the theoretical foundations for the iterative geodesic distance field refinement process.

\begin{theorem}[Geodesic Refinement Convergence]\label{thm:convergence}
Let $\mathcal{T}: \mathbb{R}^{H \times W} \rightarrow \mathbb{R}^{H \times W}$ denote the geodesic refinement operator defined by:
\begin{align}
\mathbf{U}^{(t)}_{i,j} &= \min_{(k,l) \in \mathcal{N}(i,j)} \mathbf{G}^{(t-1)}_{k,l} + \frac{\lambda}{\mathbf{F}_{i,j} + \epsilon}, \label{eq:update_term}\\
\mathbf{G}^{(t)}_{i,j} &= 
\begin{cases}
\mu \mathbf{G}^{(t-1)}_{i,j} + (1-\mu) \mathbf{U}^{(t)}_{i,j}, & \text{if } \mathbf{M}_{i,j} > \tau, \\
\mathbf{G}^{(t-1)}_{i,j}, & \text{otherwise}.
\end{cases} \label{eq:iterative_update}
\end{align}
Then $\mathcal{T}$ is a contraction mapping with Lipschitz constant $\mu < 1$, and the sequence $\{\mathbf{G}^{(t)}\}_{t=0}^{\infty}$ converges to a unique fixed point.
\end{theorem}

\vspace{0.5em}

\noindent\textbf{Proof.}
The mixing parameter $\mu = 0.7$ in Eq.~\eqref{eq:iterative_update} ensures contractivity. For any two distance fields $G_1, G_2$:
$$\|T(G_1) - T(G_2)\|_2 \leq \mu \|G_1 - G_2\|_2$$
Since $\mu < 1$, the Banach fixed-point theorem guarantees convergence.

\begin{lemma}[Numerical Stability]\label{lem:numerical_stability}
The numerical scheme in Eq.~\eqref{eq:iterative_update} is stable under the condition:
\begin{align}
\mu \leq 1 - \frac{\lambda}{\min_{i,j} \mathbf{F}_{i,j}}
\end{align}
Given our parameter choices ($\mu = 0.7$, $\lambda = 0.1$) and the speed function bounds from Lemma~\ref{lem:speed_bounds}, this condition is satisfied for all practical cases.
\end{lemma}

\begin{remark}
The regularization parameter $\epsilon = 10^{-6}$ in Eq.~\eqref{eq:update_term} ensures numerical stability by preventing division by zero while maintaining approximation accuracy to machine precision.
\end{remark}

\subsection{Prototype Space Analysis}
\label{sec:prototype_analysis}

We analyze the theoretical properties of our unified prototype representation that combines both local and global spatial information.

\begin{definition}[Unified Prototype Space]\label{def:prototype_space}
The unified prototype space $\mathbf{P}_{\text{unified}} = \{\mathbf{P}_{grid}\} \cup \{\mathbf{p}_{\text{global}}\}$ forms a complete representation of the support features with both local and global information. Specifically:
\begin{enumerate}[label=(\roman*)]
    \item \textbf{Local prototypes:} $\\mathbf{P}_{grid}\}$ capture spatial heterogeneity through adaptive grid sampling with density $\rho_{i,j} = 1 + \sigma \cdot \tau \cdot \|\nabla \mathbf{G}_{i,j}\|_2$ where $\sigma \in [1.0, 4.0]$ and $\tau \in [0.5, 3.0]$;
    \item \textbf{Global prototype:} $\mathbf{p}_{\text{global}}$ provides anatomical structure coherence via geodesic-weighted aggregation using Eq.~\ref{eq:adaptive_weighting};
    \item \textbf{Completeness:} The unified space spans both fine-grained boundary details and global anatomical patterns for comprehensive medical image representation.
\end{enumerate}
\end{definition}

\begin{lemma}[Adaptive Density Bounds]\label{lem:adaptive_density}
The adaptive sampling density $\rho_{i,j} = 1 + \sigma \cdot \tau \cdot \|\nabla \mathbf{G}_{i,j}\|_2$ with $\sigma \in [1.0, 4.0]$ and $\tau \in [0.5, 3.0]$ ensures:
\begin{align}\label{eq:adaptive_density_bounds}
1 \leq \rho_{i,j} \leq 1 + 12\max_{i,j} \|\nabla \mathbf{G}_{i,j}\|_2
\end{align}
providing bounded density variation that concentrates sampling near boundaries.
\end{lemma}

\begin{theorem}[Prototype Bounds]\label{thm:prototype_bounds}
Each grid prototype $\mathbf{P}_{g,h}$ computed via:
\begin{align}\label{eq:adaptive_grid_prototype_analysis}
\mathbf{P}_{grid} = \frac{\sum_{(i,j) \in \mathcal{N}_{g,h}} \rho_{i,j} \cdot w_{i,j} \cdot \mathbf{M}_{i,j} \cdot \mathbf{F}_{i,j}}{\sum_{(i,j) \in \mathcal{N}_{g,h}} \rho_{i,j} \cdot w_{i,j} \cdot \mathbf{M}_{i,j} + \epsilon}
\end{align}
satisfies:
\begin{align}
\|\mathbf{P}_{grid}\|_2 \leq \max_{(i,j) \in \mathcal{N}_{g,h}} \|\mathbf{F}_{i,j}\|_2
\end{align}
ensuring bounded prototype magnitudes.
\end{theorem}

\vspace{0.5em}

\noindent\textbf{Proof.}
The normalization in Eq.~\eqref{eq:adaptive_grid_prototype_analysis} ensures that $\mathbf{P}_{g,h}$ is a convex combination of feature vectors $\mathbf{F}_{i,j}$ within the neighborhood $\mathcal{N}_{g,h}$. The bound follows from the convexity of the $\ell_2$ norm.

\begin{corollary}[Prototype Convergence]\label{cor:convergence_prototype}
Under bounded feature assumptions, the prototype computation converges to a stable representation with finite norm.
\end{corollary}

\clearpage
\twocolumn
\section{Qualitative results}
\label{Qualitative results}

\begin{figure}[h]
  \centering
   \includegraphics[width=\columnwidth]{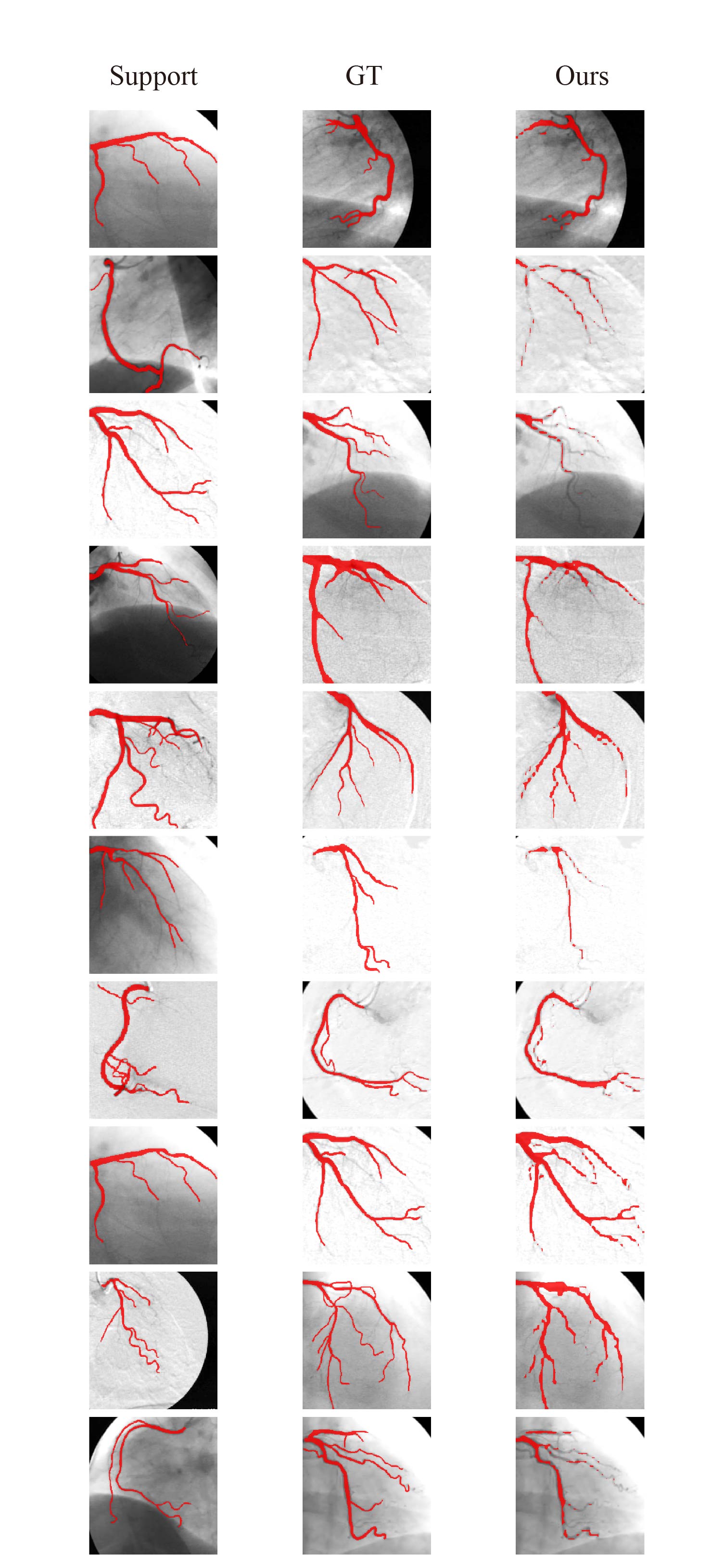}
   \caption{Visualization from left to right on DCA1 dataset. left row to right: Support, Ground-truth. Ours prediction.}
   \label{fig:DCA1-visual}
\end{figure}

\begin{figure}[t]
  \centering
   \includegraphics[width=\columnwidth]{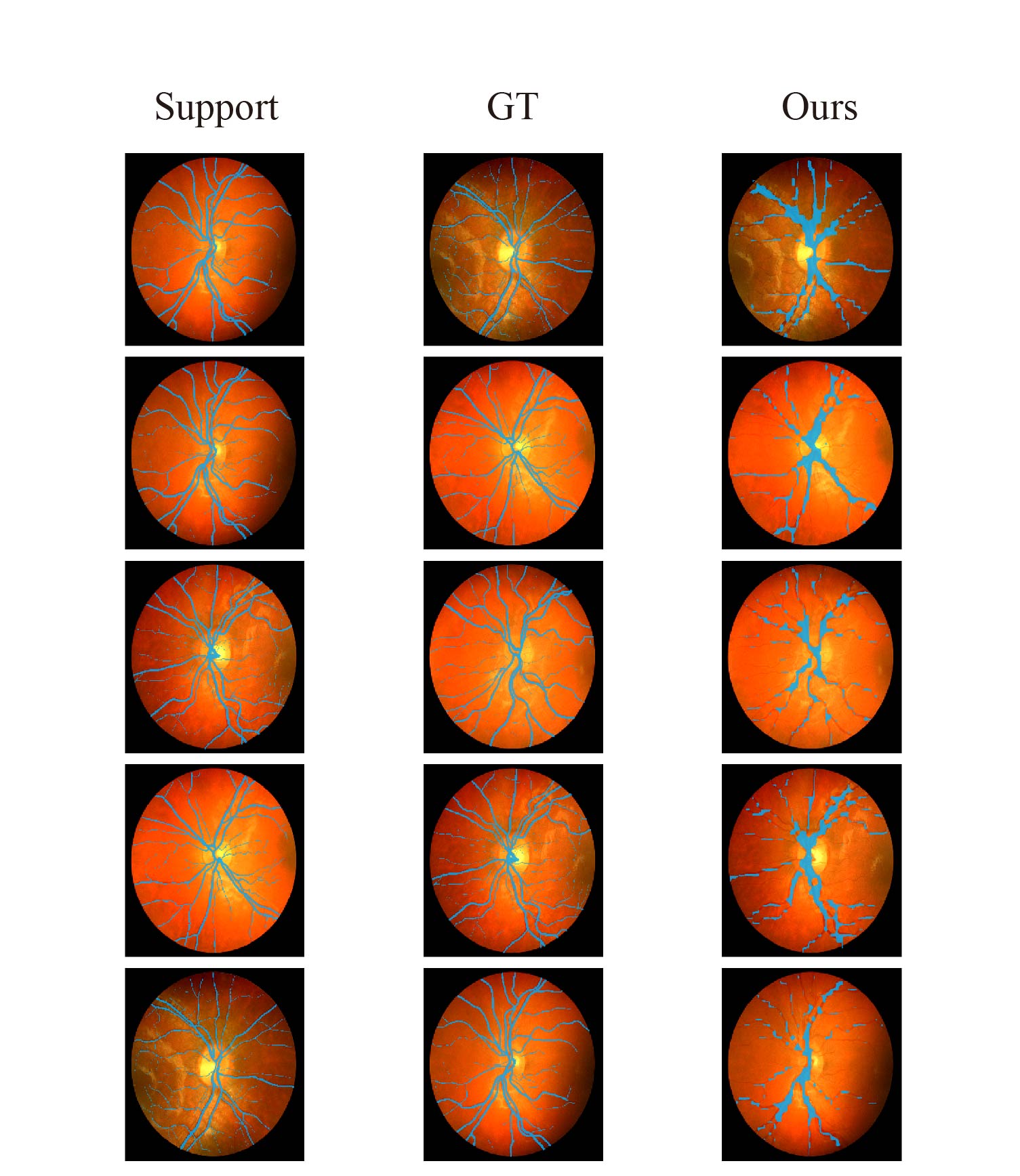}
   \caption{Visualization from left to right on CHASE-DB1 dataset. left row to right: Support, Ground-truth. Ours prediction.}
   \label{fig:chase-db1-visual}
\end{figure}

\begin{figure}[t]
  \centering
   \includegraphics[width=\columnwidth]{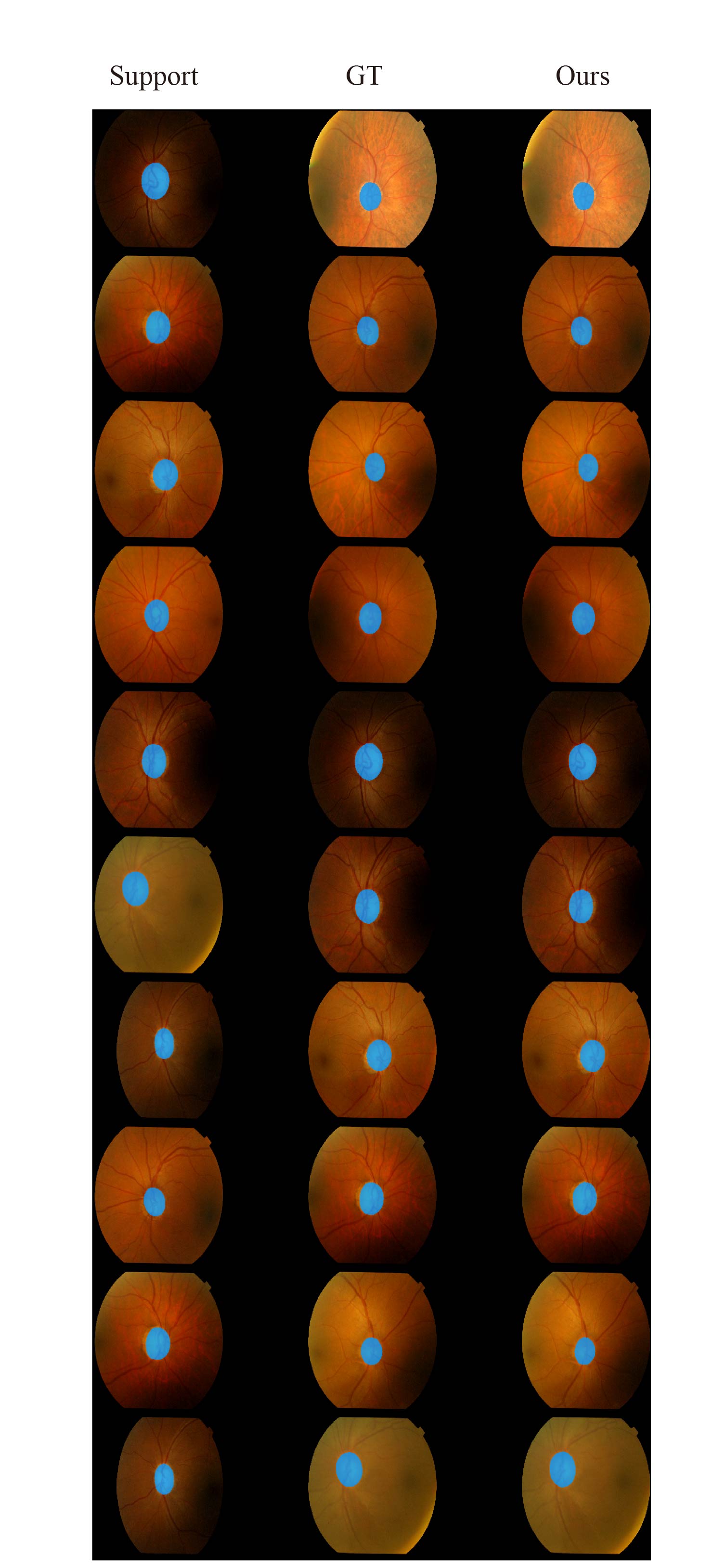}
   \caption{Visualization from left to right on Drishti-GS dataset. left row to right: Support, Ground-truth. Ours prediction.}
   \label{fig:DRISHTI-visual}
\end{figure}

\begin{figure}[t]
  \centering
   \includegraphics[width=\columnwidth]{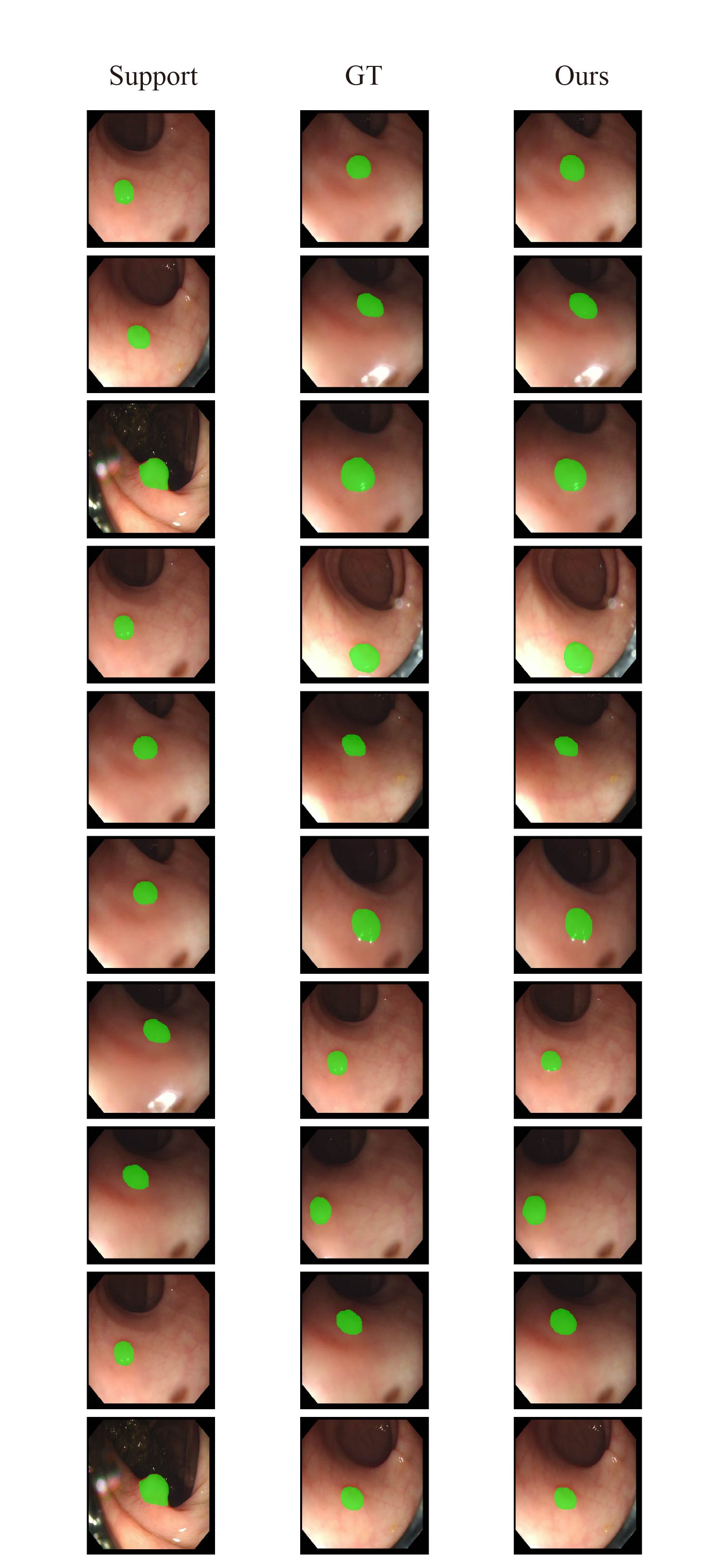}
   \caption{Visualization from left to right on CVC300 dataset. left row to right: Support, Ground-truth. Ours prediction.}
   \label{fig:cvc300-visual}
\end{figure}

\begin{figure}[t]
  \centering
   \includegraphics[width=\columnwidth]{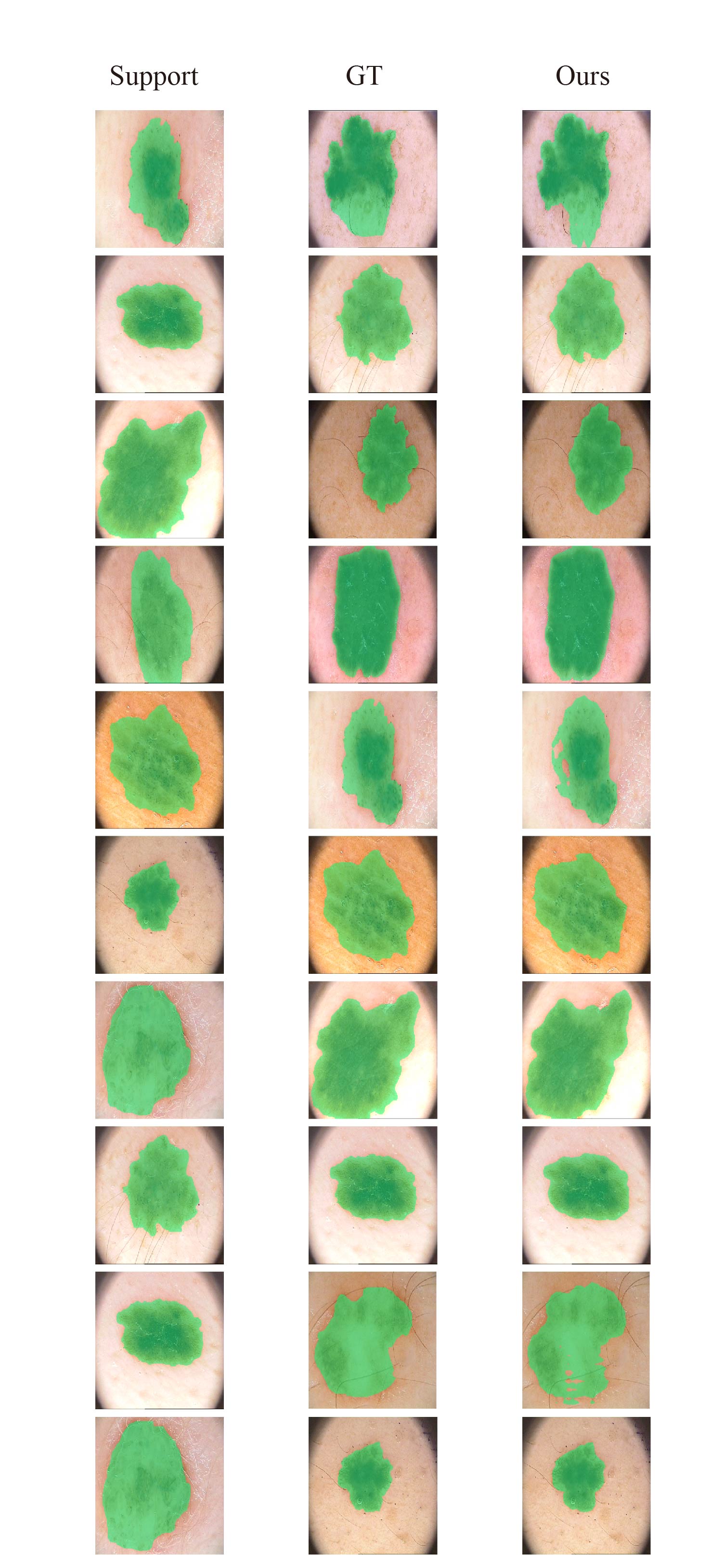}
   \caption{Visualization from left to right on PH2-typical Nevus dataset. left row to right: Support, Ground-truth. Ours prediction.}
   \label{fig:Atypical Nevus-visual}
\end{figure}

\begin{figure}[t]
  \centering
   \includegraphics[width=\columnwidth]{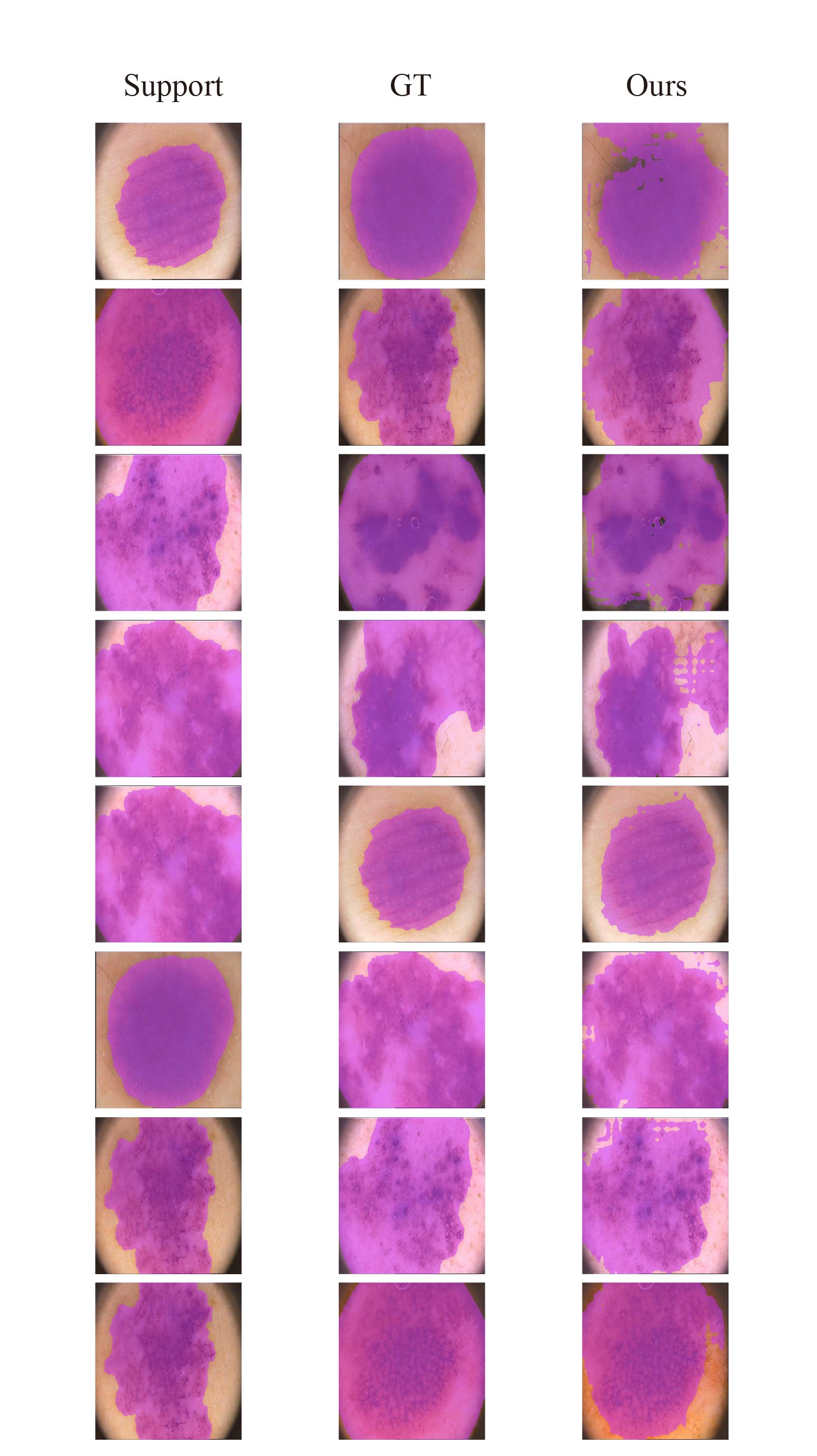}
   \caption{Visualization from left to right on PH2-Melanomas dataset. left row to right: Support, Ground-truth. Ours prediction.}
   \label{fig:_Melanomas-visual}
\end{figure}

\end{document}